\documentclass[10pt,twocolumn,letterpaper]{article}

\usepackage[pagenumbers]{cvpr}

\usepackage{graphicx}
\usepackage{amsmath}
\usepackage{amssymb}
\usepackage{booktabs}
\usepackage{color,soul}

\usepackage{caption}
\usepackage{subcaption}
\usepackage{array}
\usepackage{multirow}
\usepackage{threeparttable}

\usepackage[breaklinks=true,bookmarks=false,backref=page]{hyperref}

\usepackage[capitalize]{cleveref}
\crefname{section}{Sec.}{Secs.}
\Crefname{section}{Section}{Sections}
\Crefname{table}{Table}{Tables}
\crefname{table}{Tab.}{Tabs.}

\usepackage{enumitem}

\begin{document}

\title{Full-Body Cardiovascular Sensing with Remote Photoplethysmography}

\author{Lu Niu, Jeremy Speth, Nathan Vance, Ben Sporrer, Adam Czajka, Patrick Flynn\\
University of Notre Dame\\
{\tt\small \{lniu,jspeth,nvance1,bsporrer,aczajka,flynn\}@nd.edu}
}
\maketitle

\begin{abstract}
Remote photoplethysmography (rPPG) allows for noncontact monitoring of blood volume changes from a camera by detecting minor fluctuations in reflected light.
Prior applications of rPPG focused on face videos. In this paper we explored the feasibility of rPPG from non-face body regions such as the arms, legs, and hands. We collected a new dataset titled Multi-Site Physiological Monitoring (MSPM), which will be released with this paper.
The dataset consists of 90 frames per second video of exposed arms, legs, and face, along with 10 synchronized PPG recordings.
We performed baseline heart rate estimation experiments from non-face regions with several state-of-the-art rPPG approaches, including chrominance-based (CHROM), plane-orthogonal-to-skin (POS) and RemotePulseNet (RPNet).
To our knowledge, this is the first evaluation of the fidelity of rPPG signals simultaneously obtained from multiple regions of a human body.
Our experiments showed that skin pixels from arms, legs, and hands are all potential sources of the blood volume pulse.
The best-performing approach, POS, achieved a mean absolute error peaking at 7.11 beats per minute from non-facial body parts compared to 1.38 beats per minute from the face. Additionally, we performed experiments on pulse transit time (PTT) from both the contact PPG and rPPG signals. We found that remote PTT is possible with moderately high frame rate video when distal locations on the body are visible. These findings and the supporting dataset should facilitate new research on non-face rPPG and monitoring blood flow dynamics over the whole body with a camera.
\end{abstract}
\section{Introduction}

Heart rate is one of the most important vital signs. Photoplethysmography (PPG)~\cite{Allen2007} is an optical technique to detect light changes caused by blood flow to noninvasively estimate heart rate. Traditional PPG employs a sensor that transmits and measures reflected light of specific frequencies on vascularized skin of the fingertip, ears, or forehead. These sensors are affixed to the body, which can be inconvenient and can present risks to neonates, the elderly, and patients with damaged skin~\cite{Zhan2020-hm, informatics9030057, 6226654, DeHaan2013}. These concerns, plus increased interest in image-based noncontact measurement of vital signs, have yielded increasing interest in remote photoplethysmography (rPPG), which can estimate blood flow changes at a distance, including such exotic applications as measurement from UAV-mounted cameras~\cite{Huang_ICCRE_2020}.

The current major databases conducive to rPPG are mainly focused on facial skin pixels. MAHNOB-HCI~\cite{5975141} and PURE~\cite{Stricker2014} contain face videos with limited or controlled head movements. MMSE-HR~\cite{Zhang_2016_CVPR} collected face videos after inducing emotional responses. UBFC-rPPG~\cite{Bobbia2019} contains occasional movements and subjects under stress from mathematical games. DDPM~\cite{Speth_IJCB_2021, Vance2022} was the first long-form dataset containing unconstrained facial movements. Synthetic face rPPG datasets containing avatars also only contain faces~\cite{mcduff2022scamps,Kadambi2022}. As rPPG becomes more common for ubiquitous health monitoring, it is important to understand the limitations presented by partial or total occlusion of the face, as well as the corresponding opportunities when skin regions {\em not} on the face are visible. 

\begin{figure*}
    \centering
    \includegraphics[width=\linewidth]{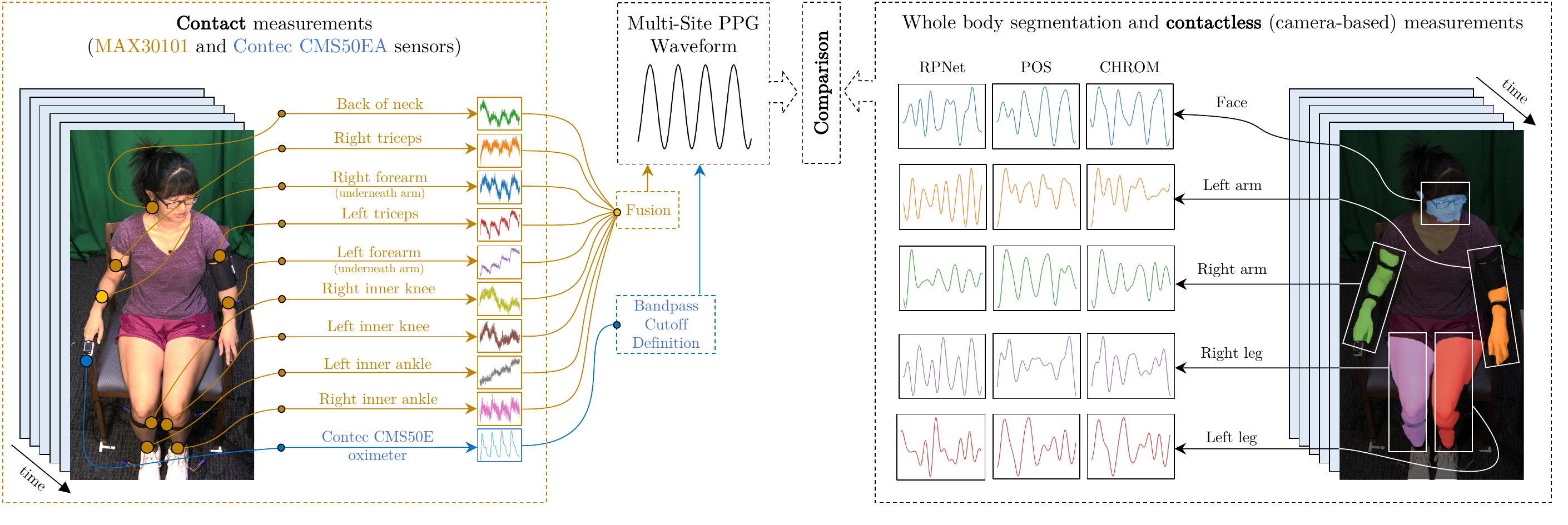}
    \caption{This paper explored the (a) Multi-Site Physiological Monitoring (MSPM) dataset with {\bf contact} measurements from ten PPG sensors, including a fingertip oximeter, as shown on the left, with (b) {\bf contactless} (camera-based) measurements of multiple skin areas carried out by three state-of-the-art rPPG methods (CHROM~\cite{DeHaan2013}, POS~\cite{Wang2017}, and RPNet~\cite{Speth_CVIU_2021}) as shown on the right. In this paper we also compared de-noised and combined PPG signals with the corresponding rPPG estimates.}
    \label{fig:teaser}
\end{figure*}

This paper explores multi-site rPPG, where the blood volume pulse is simultaneously estimated from the face, arms, hands, and legs. From a physiological perspective, collecting rPPG signals from different skin regions could provide a richer map of blood transport through the skin's capillary networks, rather than a scalar measurement of periodic blood volume changes at a single location~\cite{Allen2007, Mukkamala2015, Jeong2016, Bentham2018-fd, Chan2019-mi, Block2020, Iuchi_2022_CVPR, Iuchi_BOE_2022}. Figure~\ref{fig:teaser} displays the experimental process presented in this paper. Our main contributions are summarized as follows:

\begin{enumerate}[label=(\alph*)]
    \item The newly-created Multi-Site Physiological Monitoring (MSPM) dataset collected from eighty-seven subjects. This dataset is, to our knowledge, the first dataset that allows simultaneous rPPG estimation from multiple body sites including arms, legs, and hands in addition to the face (see Sec. \ref{sec:dataset}); 
    \item A comprehensive evaluation of baseline rPPG methods applied to video from multiple skin regions of the body. We used state-of-the-art rPPG methods including CHROM~\cite{DeHaan2013}, POS\cite{Wang2017}, and RPNet~\cite{Speth_CVIU_2021} to retrieve signals from each body part, as well as sub-regions of each body part, to explore the overall performance and motivate future work based on these results (see Sec. \ref{sec:approach}, \ref{sec:local} and \ref{sec:global}).
    \item A pulse transit time (PTT) analysis from both contact PPG and rPPG measurements at different sites on the body. We showed that pulse transit time can be estimated from a camera with lower frame rate than previous studies~\cite{Jeong2016, Iuchi_BOE_2022} if the measurement sites on the body are further apart (see Sec. \ref{sec:ptt}).
\end{enumerate}

\section{Related Work}

\subsection{Camera-based Measurements}
rPPG methods estimate pulse rate based on skin pixel color changes from blood volume fluctuations. Several rPPG algorithms have been proposed based on different mechanisms, including Blind source separation (BSS)~\cite{Wedekind_2017} and independent component analysis (ICA)~\cite{Poh2010, Poh2011}, which estimates pulse rate by applying different criteria to separate temporal RGB traces into uncorrelated or independent signal sources~\cite{Wang2017}. CHROM~\cite{DeHaan2013} assumes a standardized skin tone under white light and linearly combines the color signals for heart rate estimation. Spatial subspace rotation (2SR)~\cite{Wang2016} utilizes both spatial subspace and temporal rotation angle to calculate heart rate. POS~\cite{Wang2017} applies a plane orthogonal to the skin tone in the temporally normalized RGB space for pulse extraction.

More recently, deep learning models have been developed as the size of rPPG databases has increased. The first deep learning method~\cite{Hsu2014} used a support vector regression model on both ICA and chrominance features. Convolutional neural networks (CNN) have successfully been applied to frame differences with several adaptations~\cite{Chen2018, Liu_MTTS_2020, Liu_2023_WACV}. End-to-end waveform estimation from cropped videos passed through 3D-CNNs was first presented with PhysNet~\cite{Yu2019} and later improved with internal dilated convolutions to improve temporal context~\cite{Speth_CVIU_2021}. Several unsupervised approaches have been introduced to reduce the need for simultaneous PPG ground truth. Most approaches leverage contrastive training strategies, where similar pairs of samples are pulled closer and different samples are repelled~\cite{Gideon_2021_ICCV, Wang_SSL_2022, Yuzhe_SimPer_2022, Sun_2022_ECCV}. The first non-contrastive approach leverages strong periodic priors to encourage the model to predict sparse signals in the frequency domain~\cite{speth2023sinc}.

\subsection{Multiple Contact Measurements}
Measuring contact PPG from multiple locations simultaneously has garnered research interest primarily for blood pressure estimation from features related to pulse transit time (PTT)~\cite{Allen2007,Mukkamala2015,RibasRipoll2019,Block2020,Lubin2021,Natarajan2022}. Unfortunately, such systems are cumbersome, since multiple sensors must be properly attached and synchronized. It is therefore desirable to gather the same physiological information from a single sensor without contact, thus providing additional motivation for this work. Another advantage is that the number of potential transit time differences is drastically increased when using a camera sensor over contact PPG.

\begin{figure*}
        \centering
        \includegraphics[width=0.85\linewidth]{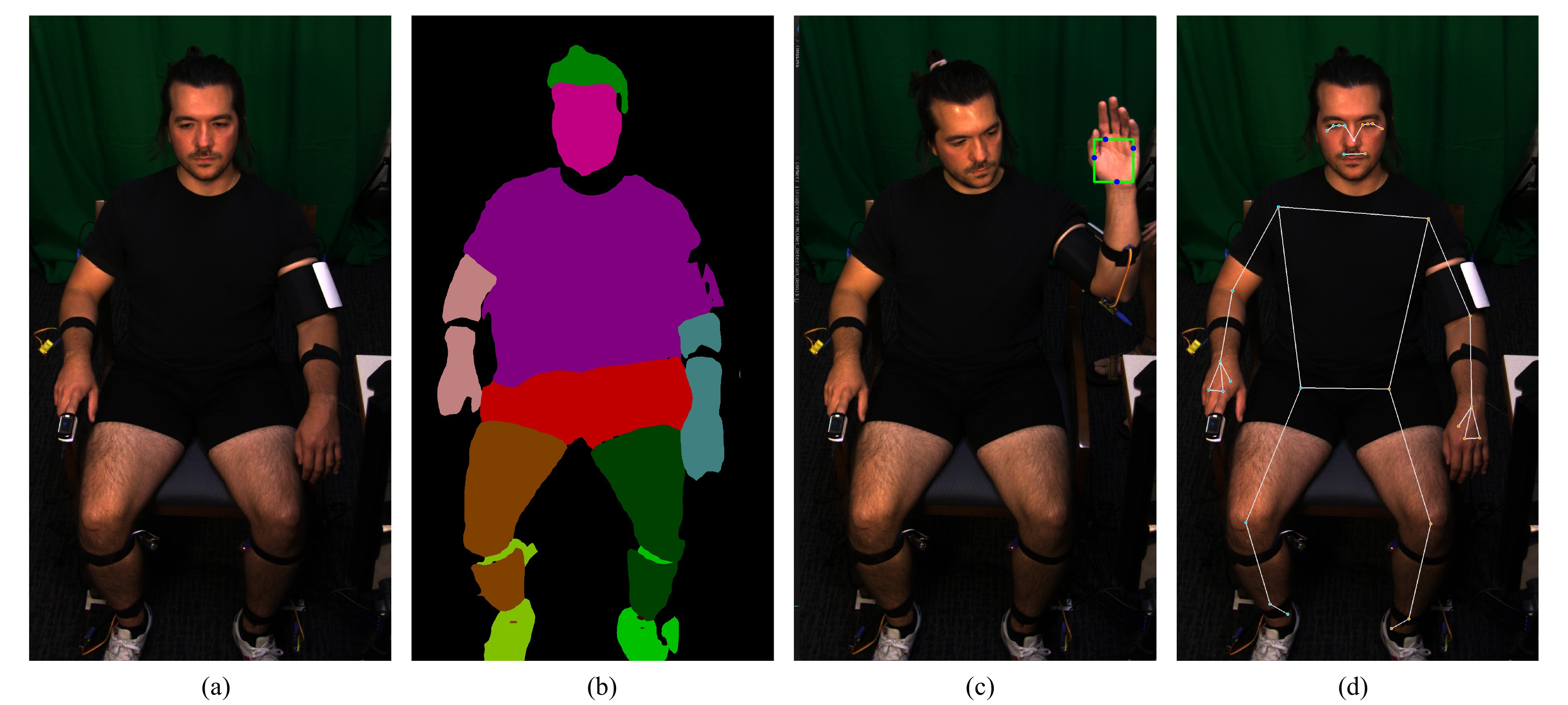}
        \caption{The subjects remained seated while collecting data (a). Using a trained SCHP~\cite{li2020self} model on the LIP~\cite{https://doi.org/10.48550/arxiv.1804.01984} dataset, the human body was segmented into 20 classes (b). rPPG Signals were extracted separately from 5 classes including 'Face', 'Left-arm', 'Right-arm', 'Left-leg', 'Right-leg'. We used Mediapipe Hand detection~\cite{Zhang2020MPHands} to detect 4 points (shown in blue) of palm and use them to form a bounding box (shown in green) (c). Mediapipe Pose detection method~\cite{Bazarevsky2020} was applied and thirty-three keypoints of human body were generated (d).}
        \label{fig:frames}
\end{figure*}

\subsection{Non-Facial rPPG}
Most of the existing research utilizes the diffuse reflection from the face region to estimate the rPPG signal, with the exception of a few works analyzing the hand~\cite{Kamshilin2011, Teplov2014, Saiko2021, Cao2023}, arms~\cite{Zaytsev2018}, and thigh~\cite{Harford2022}. A gap in current rPPG research is the capture and estimation at multiple parts of the body simultaneously. Understanding rPPG signal quality across the entire body could allow for fewer constraints on the user and create new biomarkers as blood volume dynamics are measured in the peripheral vasculature.

\section{Multi-Site Physiological Monitoring Dataset}
\label{sec:dataset}

We collected a large dataset of subjects seated in front of a video camera with several attached contact PPG sensors. Subjects were instructed to avoid wearing clothes that obstructed the arms and legs. A sample video frame is shown in Fig.~\ref{fig:teaser}, where the subject's face, arms, and legs are visible simultaneously. To the authors' knowledge, this is the first dataset that allows for simultaneous ground truthed rPPG at more than two sites on the body.

The collection began with the subject holding their left hand upright facing the camera for 90 seconds. This allows for evaluation of rPPG on the palm, a region with glabrous skin that is valuable for remote pulse estimation~\cite{Cao2023}. For the remainder of the session, subjects underwent a blood pressure measurement, guided breathing exercise, and relaxation period. On average, the sessions evaluated in this paper last for approximately 5.7 minutes.

\subsection{Apparatus}
Video of the full subject was recorded with a DFK 33UX290 RGB camera from The Imaging Source (TIS). Raw video was recorded at 90 FPS with a resolution of 1920 $\times$ 1080 pixels. To record blood oxygenation (SpO2) and pulse rate, we collected 60 samples per second from an FDA-certified Contec CMS50EA pulse oximeter attached to an index finger. To support research in pulse transit time, we attached nine MAX30101 contact PPG sensors that recorded red and near-infrared signals at 400 samples per second. Each leg and arm had two attached sensors via elastic straps, and the last sensor was attached to the back of the neck with medical tape. All sensors and the camera recorded raw data to a single collection machine with the associated timestamps for easy synchronization.

\subsection{Data Preprocessing}
To properly evaluate rPPG approaches, we created a global pulse rate estimate from the multiple contact sensors. Contact PPG signals recorded from the nine sites with the MAX30101 sensors contain varying levels of noise throughout the session depending on body movements. However, the pulse signal is likely present in at least one signal at any given point in time, since movements may be isolated to local regions.

Using this assumption, we combined the multiple pulse signals using a sliding window approach and bandpass filtering. We relied on the FDA-certified CMS50EA oximeter's pulse rate estimate to define the bounds of a narrow bandpass filter for the MAX30101 signals. Given the estimated pulse rate from the fingertip oximeter $Y$, we specified a padding around this value, $\Delta Y = 30$ bpm, and filtered the MAX30101 signals with a 2nd order Butterworth filter with lower and upper cutoffs of $Y - \Delta Y$ and $Y + \Delta Y$, respectively.

\begin{figure}
    \centering
    \includegraphics[width=\linewidth]{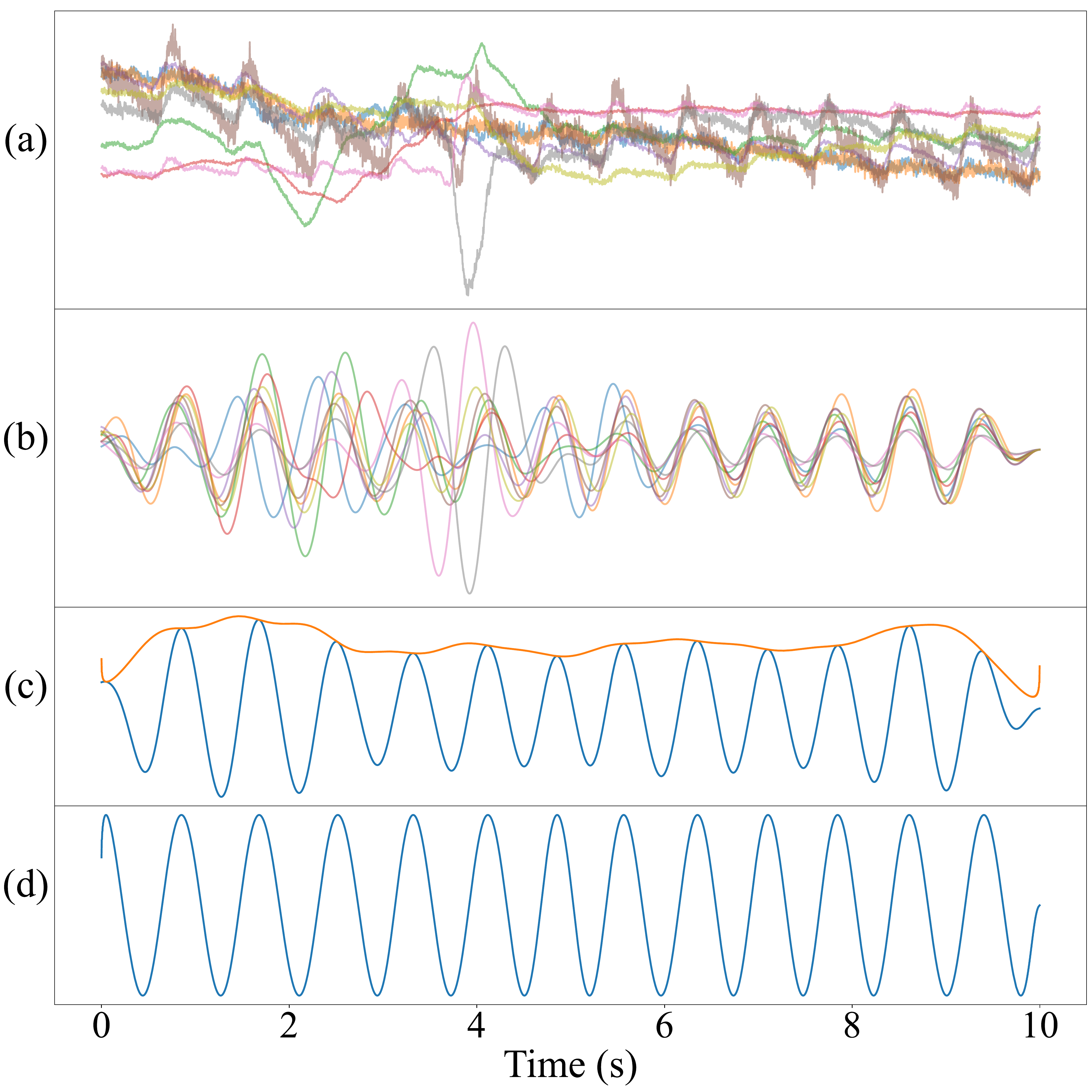}
    \caption{Processing of the contact PPG waveforms to produce a robust pulse rate. (a) The z-normalized waveforms from all sensors. (b) Bandpass filtered signals around the fingertip oximeter's pulse rate $\pm$ 30 bpm. (c) The signals were added together and the envelope is calculated. (d) The combined signal was divided by the envelope.}
    \label{fig:ground_truth}
\end{figure}

Specifically, for a sliding 10-second window with a stride of a single sample at 400 Hz, the waveforms underwent z-normalization, followed by filtering around the fingertip oximeter's pulse rate, then they were summed together into the combined waveform for that window. Finally, we divided the complete combined waveform by its envelope, which was calculated from the Hilbert transform. The process for signal combination is shown in Fig.~\ref{fig:ground_truth}. In the figure, between seconds 2 and 5, there appears to be noise from motion in multiple sensors, but the combined signal in subplot (d) remains clean.

\section{Multi-Site rPPG Approach}
\label{sec:approach}
Our pipeline to perform remote pulse estimation on our MSPM dataset consists of preprocessing, signal extraction and pulse rate estimation. We applied three pulse estimation algorithms: a chrominace-based method using color difference signals (CHROM)~\cite{DeHaan2013}, plane-orthogonal-to-skin tone (POS)~\cite{Wang2017}, and a 3D-CNN deep learning method RPNet~\cite{Speth_CVIU_2021}. These three methods perform well for both stationary and moving subjects, which are characteristic of our MSPM dataset. The RPNet model was trained on face videos from the large and challenging deception detection and physiological monitoring dataset (DDPM)~\cite{Speth_IJCB_2021,Vance2022}. The model was not fine-tuned on any data from MSPM, so we tested how well the model could transfer to new subjects, lighting, standoff, and movement. Since the model was only trained on cropped faces, we also investigated its ability to transfer to spatially dissimilar regions such as the arms and legs. One of the benefits of CHROM and POS is that they do not have any spatial priors and only expect RGB traces from skin pixels.

\subsection{Preprocessing}
\subsubsection{ROI selection}\label{sec:ROI}
Given skin pixels including face, arms, and legs collected by MSPM, we selected these five different body parts as our ROIs. We applied Self-Correction Human Parsing (SCHP)~\cite{li2020self} for body segmentation to the eighty-seven subjects of MSPM. Frame (b) of Fig.~\ref{fig:frames} shows masks of the twenty parts from SCHP for a single frame. Masked skin pixels of each ROI were averaged and supplied to CHROM and POS for signal extractions. For each masked ROI, we also calculated the minimum and maximum values of $(x, y)$ locations of each mask to form bounding boxes for RPNet. Similarly, we created bounding boxes for palms using four key points of the palm generated by Mediapipe Hand detection~\cite{Zhang2020MPHands}. Frame (c) of Fig.~\ref{fig:frames} shows the bounding box for the palm.

\subsubsection{Human Key Points Detection}
To better investigate rPPG performance, we applied Mediapipe Pose detection~\cite{Bazarevsky2020} to detect key points of the human skeleton for local error estimation of the whole body (described later in Sec. \ref{sec:local}). There are thirty-three key points in total which define locations of many of the body's joints, including detailed locations for hands. 
Frame (d) of Fig.~\ref{fig:frames} shows an example of the skeleton detection results.

\subsection{Signal Extraction}
\subsubsection{Color Transformation Methods}

For each body part, we calculated spatial averages of skin pixels to reduce camera quantization error and generate 1D signals for each channel of RGB. CHROM~\cite{DeHaan2013} and POS~\cite{Wang2017} combined the signals of the three channels linearly to remove noise from movement or lighting to generate more robust pulse signals.

\setlength\tabcolsep{3.9pt}
\begin{table*}[htb!]
\caption{Pulse rate estimation results. MAE: Mean Absolute Error; $r$: Pearson correlation coefficient.}
\fontsize{9.0}{9}\selectfont
\centering
\begin{tabular}{l cc cc cc cc cc | cc | cccc}
\toprule
& \multicolumn{10}{c|}{Both relaxed and hand raise} & \multicolumn{2}{c|}{Relaxed} & \multicolumn{4}{c}{Hand raise}\\
\cmidrule(lr){2-11}\cmidrule(lr){12-13}\cmidrule(lr){14-17}
\multirow{3}{*}{\begin{tabular}[c]{@{}l@{}}\\Methods\end{tabular}} &
\multicolumn{2}{c}{Face} &
\multicolumn{2}{c}{Right leg} &
\multicolumn{2}{c}{Left leg} &
\multicolumn{2}{c}{Right arm} &
\multicolumn{2}{c|}{Left arm} &
\multicolumn{2}{c|}{Left arm} &
\multicolumn{2}{c}{Left arm} &
\multicolumn{2}{c}{Palm}\\ 
\cmidrule(lr){2-3}
\cmidrule(lr){4-5}
\cmidrule(lr){6-7}
\cmidrule(lr){8-9}
\cmidrule(lr){10-11}
\cmidrule(lr){12-13}
\cmidrule(lr){14-15}
\cmidrule(lr){16-17}
& \begin{tabular}[c]{@{}c@{}}MAE\\ (bpm)\end{tabular} & $r$
& \begin{tabular}[c]{@{}c@{}}MAE\\ (bpm)\end{tabular} & $r$
& \begin{tabular}[c]{@{}c@{}}MAE\\ (bpm)\end{tabular} & $r$
& \begin{tabular}[c]{@{}c@{}}MAE\\ (bpm)\end{tabular} & $r$
& \begin{tabular}[c]{@{}c@{}}MAE\\ (bpm)\end{tabular} & $r$
& \begin{tabular}[c]{@{}c@{}}MAE\\ (bpm)\end{tabular} & $r$
& \begin{tabular}[c]{@{}c@{}}MAE\\ (bpm)\end{tabular} & $r$
& \begin{tabular}[c]{@{}c@{}}MAE\\ (bpm)\end{tabular} & $r$\\
\midrule
CHROM \cite{DeHaan2013} &
    2.38  & 0.85 &
    10.92 & 0.42 &
    11.07 & 0.41 &
    9.13  & 0.50 &
    9.81  & 0.41 &
    11.57 & 0.35 &
    4.26  & 0.71 &
    5.01  & 0.67 \\
POS \cite{Wang2017} &
    \textbf{1.38} & \textbf{0.93} &
    \textbf{6.96} & \textbf{0.54} &
    \textbf{7.11} & \textbf{0.54} &
    \textbf{3.60} & \textbf{0.78} &
    \textbf{6.04} & \textbf{0.64} &
    \textbf{6.88} & \textbf{0.61} &
    \textbf{3.40} & \textbf{0.75} &
    \textbf{3.88} & \textbf{0.76} \\
RPNet \cite{Speth_CVIU_2021}\tnote{\textdagger} &
    2.27  & 0.87 &
    29.50 & 0.14 &
    30.42 & 0.11 &
    23.94 & 0.15 &
    23.15 & 0.16 &
    27.01 & 0.11 &
    11.06 & 0.38 &
    6.70  & 0.52 \\
\bottomrule
\end{tabular}
\label{tab:within_dataset}
\end{table*}

\subsubsection{Learning-Based Method}
We used the generated bounding box coordinates to crop ROIs from each frame for each body part and downsized these ROIs to 64x64 pixels using bicubic interpolation. The RPNet model we used was trained on the DDPM dataset~\cite{Speth_IJCB_2021,Vance2022} where the frame rate is 90 frames per second (fps), which is the same as our MSPM dataset. The trained RPNet model was fed video clips of 135 frames (1.5 seconds) as described in the original paper~\cite{Speth_CVIU_2021}.

\subsection{Filtering and Pulse Rate Calculation}
Remote pulse estimation over non-face regions is challenging, because of their lower signal-to-noise ratio than the face. To improve signal quality for all approaches, we applied a 4th order Butterworth bandpass filter with cutoff frequencies of 40 bpm and 180 bpm. Bandpass filtering was not used in the original POS and RPNet implementations, but we found the POS estimates in particular to contain high frequency noise before filtering. To compute the dominant pulsatile component in the rPPG waveforms, we applied the short-time Fourier transform (STFT) over a sliding 10-second window (900 frames in our videos), and selected the spectral peak as the pulse rate.

\section{Global rPPG Experiments}
\label{sec:global}
\subsection{Global Evaluation}
We refer to rPPG with all pixels in a region of interest as ``global" rPPG. We evaluated global rPPG performance over the face, arms, and legs with the color transformation approaches (CHROM~\cite{DeHaan2013} and POS~\cite{Wang2017}), and a 3D-CNN approach (RPNet~\cite{Speth_CVIU_2021}). We compared the rPPG quality by evaluating the pulse rate performance. We applied the same method for pulse rate estimation on both the ground truth signals and the rPPG signals for fair evaluation~\cite{Mironenko2020}. 
We used popular error metrics to compare the pulse rate estimates, including mean absolute error (MAE) and Pearson's correlation coefficient ($r$).

\subsection{Global Results}

The results of global rPPG experiments with different body parts are shown in Table~\ref{tab:within_dataset}. We show results for the entire sequence as well as for separate components in which the left arm is either relaxed or raised with left palm frontally presented. We observed the best MAE and $r$ for the face region. CHROM and RPNet give similar performance on the face region, with RPNet giving slightly better performance. RPNet performed poorly on the non-face regions, however, likely indicating that the model has learned spatial priors to look for facial features. Additionally, the deep learning model may be overfit to the skin thickness, melanin concentration, and microcirculation present in the glabrous skin of the face. The improved performance for RPNet on the palm (also with glabrous skin) helps justify this explanation.

The POS algorithm gives the best performance for all body regions in terms of both MAE and $r$. This is especially impressive given that it is a simple linear method. It also shows that color changes from blood volume are similar over different skin thicknesses and underlying microvasculature. For POS and CHROM, the order of performance from best to worst is face, arms, then legs. The palm also gives good performance for all approaches, due to physical similarities to the skin on the face.

\begin{figure}
    \centering
    \includegraphics[width=\linewidth]{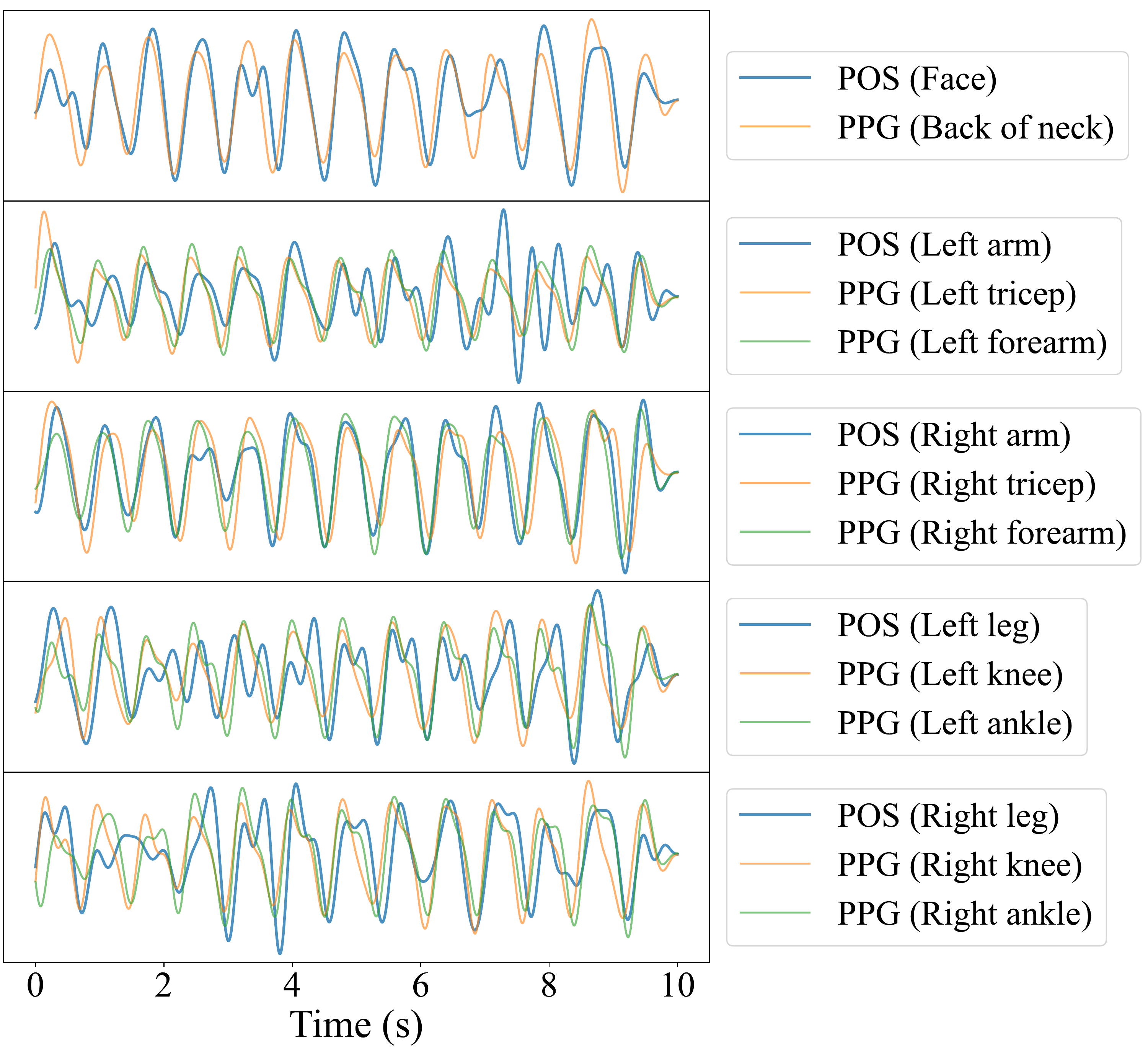}
    \caption{Pulse waveform predictions from POS overlayed with the nearest contact-PPG waveforms for a 10 second window.}
    \label{fig:result_waveforms}
\end{figure}

Figure \ref{fig:result_waveforms} shows predicted waveforms from a 10-second window for the same subject at different locations. The nearest contact PPG waveforms are bandpass filtered and plotted against the predictions to show that many of the predictions contain the underlying dominant pulse even in the presence of higher frequency noise. The difference in waveform morphology across body location shows how informative full-body rPPG can be. For this particular segment, the legs contain a strong second harmonic, which may arise from either the closure of the aortic valve during forward wave propagation, or wave reflections occurring at structural discontinuities along the femoral arteries~\cite{London1999}. Future studies on this dataset will explore different waveform morphologies at a finer scale and their relation to arterial stiffness and blood pressure.

\begin{figure}
    \centering
    \includegraphics[width=\linewidth]{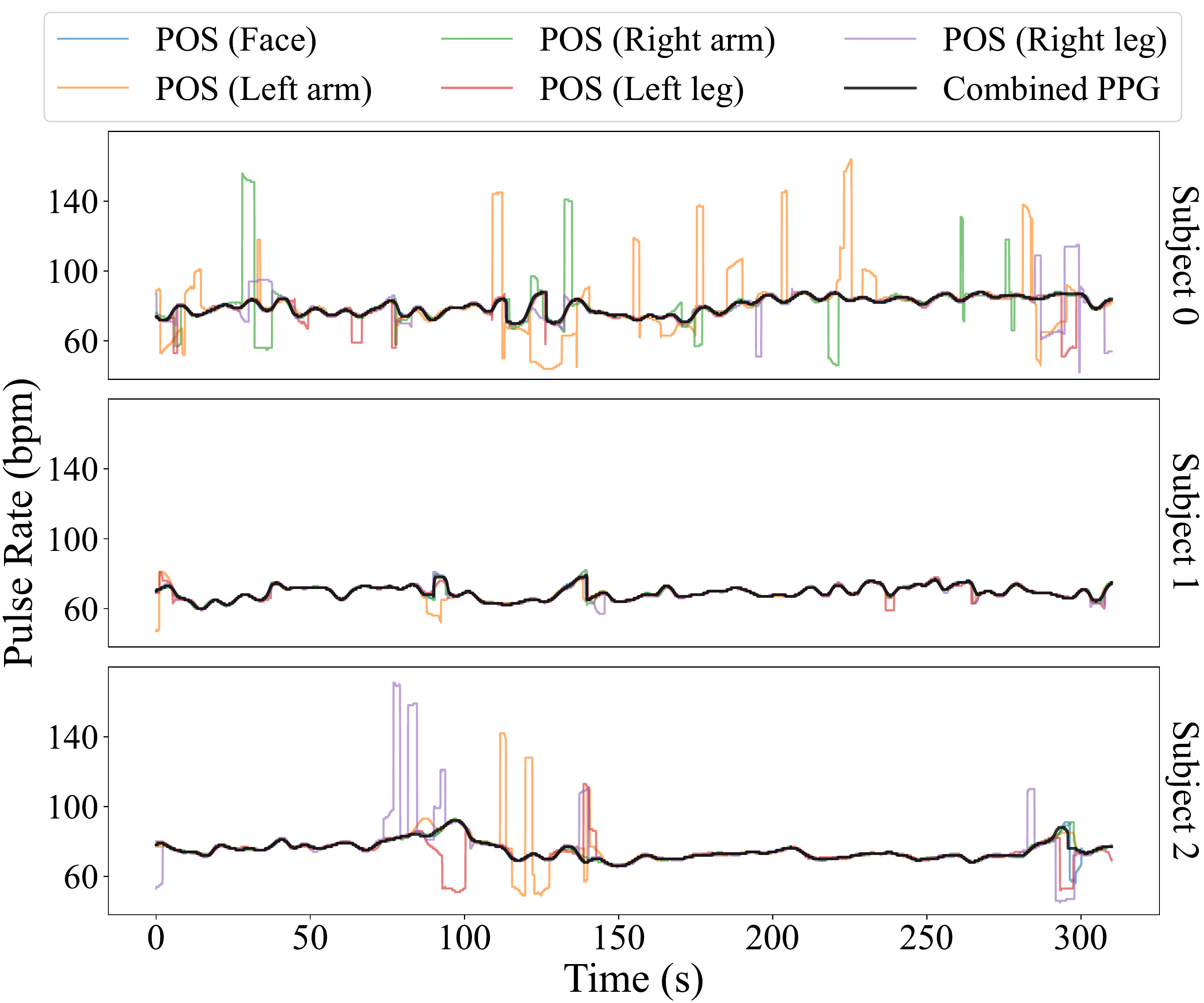}
    \caption{Pulse rate predictions from POS for 3 subjects over the course of a full session compared to the ground truth.}
    \label{fig:result_HRs}
\end{figure}

Figure~\ref{fig:result_HRs} shows pulse rates calculated from POS predictions over the entire session for 3 subjects. Interestingly, the errors generally occur as spikes of short duration rather than sustained periods of large offset. Even for the left arm, which is the worst-performing region for subject 0, the errors appear to be due to short periods where the second harmonic contains more power than the first harmonic. It is possible that simple heuristics during postprocessing could remove these transient errors. In general, the global POS signals give meaningful predictions for most applications, even in non-face regions.

\section{Local rPPG Experiments}
\label{sec:local}

\subsection{Local Evaluation}
In addition to analyzing performance for the entire face, arm, and legs, we performed a ``local" analysis on subregions of pixels in the native video resolution. This gives a more detailed evaluation on rPPG quality over the visible skin on the body. We selected POS~\cite{Wang2017} as the rPPG method, since it was the best performing for each of the global regions.

We used the ROIs described in section~\ref{sec:ROI} to define bounds for a non-overlapping sliding window of 20x20 pixel subregions. The POS algorithm was applied on 10-second non-overlapping time windows of the spatially averaged subregions for the whole video. The predictions were then linearly interpolated up to the native image resolution (i.e. 20 times as large along the x- and y-axes). Next, we computed error metrics at each pixel location for the 10-second windows, which results in 9 error frames for the hand raising portion, and around 25 error frames for the sitting portion per subject.

For error metrics we used MAE between the predicted and ground truth pulse rate and signal-to-noise ratio (SNR). The SNR was calculated similarly to \cite{DeHaan2013,Nowara_BOE_2021}, where the sum of spectral power around the signal bands in the first and second harmonics ($\pm 6$ bpm) was divided by the total power outside the signal bands.

To remove spurious local errors outside of the skin pixel regions, we applied the average SCHP mask (see frame (b) in Fig.~\ref{fig:frames}) for the time window to the error frames. Since subjects sat in slightly different positions throughout the interview, we aligned each body part across subjects for an accurate physiological error map. To do this we used the pose keypoints from Mediapipe (see frame (d) in Fig.~\ref{fig:frames}). We first calculated the average pose across all subjects for the hand raising and relaxed portions as the target poses. Then for each error frame for every subject, we applied a homographic transformation from the time window's pose to the hand raising or relaxed portion's target pose.

\begin{figure*}
    \centering
    \includegraphics[width=0.85\linewidth]{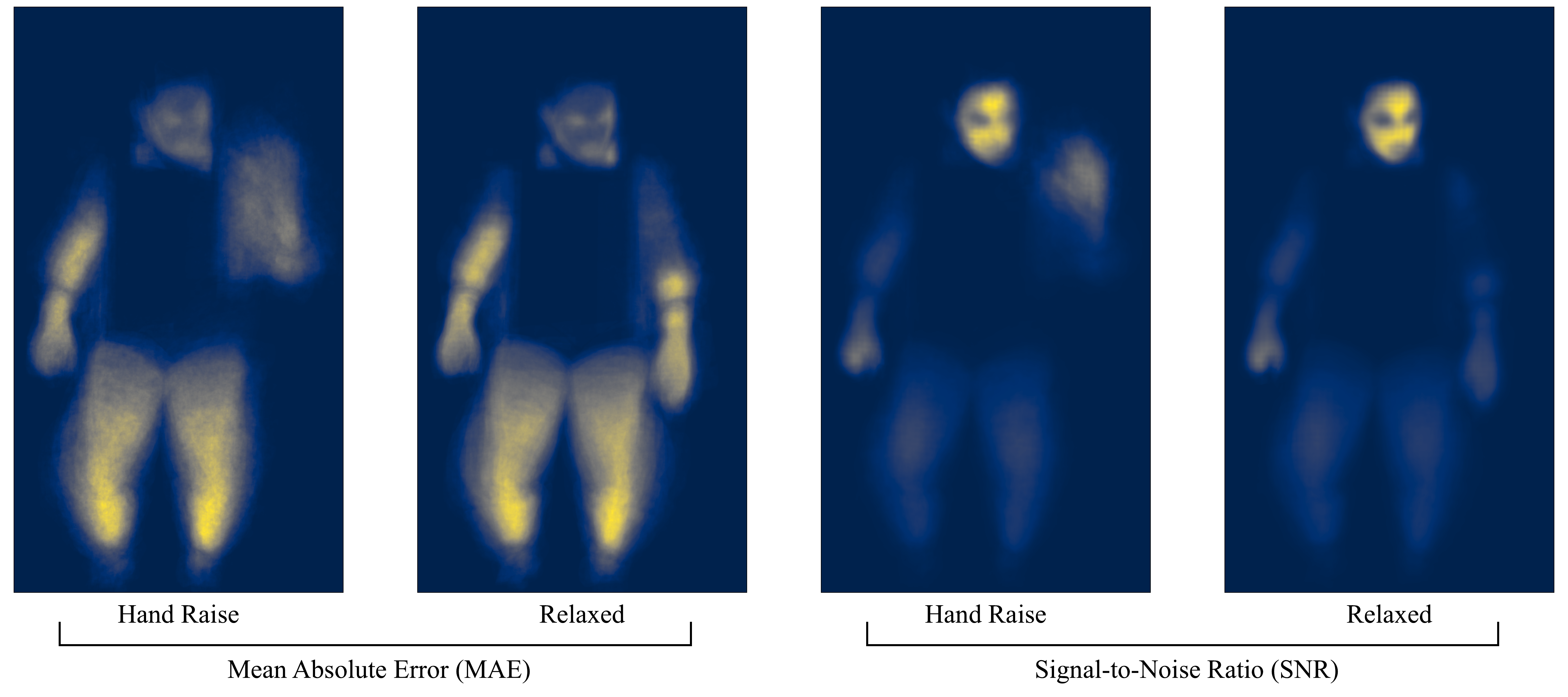}
    \caption{A heatmap of the local POS signal quality across the visible skin. POS signals are predicted over 20x20 pixel subregions, masked with SCHP~\cite{li2020self}, and the human pose is aligned with a homographic transformation.}
    \label{fig:heatmap}
\end{figure*}

\subsection{Local Results}
Figure~\ref{fig:heatmap} shows a heatmap of local pulse estimation performance for the POS algorithm. The hand raising and sitting portions are visualized separately, since the body positions were much different. In general, we found the face to hold the strongest rPPG signal. In the hand raising heatmap, we can see that the palm gave relatively low errors compared with the arms and legs. This supports the hypothesis that signal quality is higher on glabrous than non-glabrous skin~\cite{Cao2023}, and shows promise for substituting the face region with the hand if necessary.

It is useful to analyze each body part separately to assess the signal quality. Firstly, the face appears to give low MAE and high SNR in all regions except the eyes and mouth. This is in line with past research that mainly utilizes the forehead and cheek regions. The arms give relatively low signal quality, but there is a slight improvement visible above the forearm over other regions in the SNR plots. The legs give perhaps the weakest rPPG signal as evidenced by the global results in Table \ref{tab:within_dataset}. Within the local analysis, we can see that the thighs give higher SNR than the shins. This could be due to the thigh's closer proximity to the illuminators, whereas the shins are visibly darker in the video.

In both the hand raising and relaxed portions, the hand gives the second best SNR. In nearly all cases the subject's right hand was resting with the palm facing downwards, indicating that the back of the fingers and hand on non-glabrous skin is still feasible for rPPG. During the hand raising, we see that the signal quality is high on the glabrous skin of the palm. With more sophisticated methods for aligning each subject's palms, we believe the error maps would give even higher signal quality.
\section{Pulse Transit Time Experiments}
\label{sec:ptt}

\begin{figure*}
        \centering
        \begin{minipage}[t]{0.37\linewidth}
        \begin{subfigure}{\linewidth}
        \includegraphics[width=\linewidth]{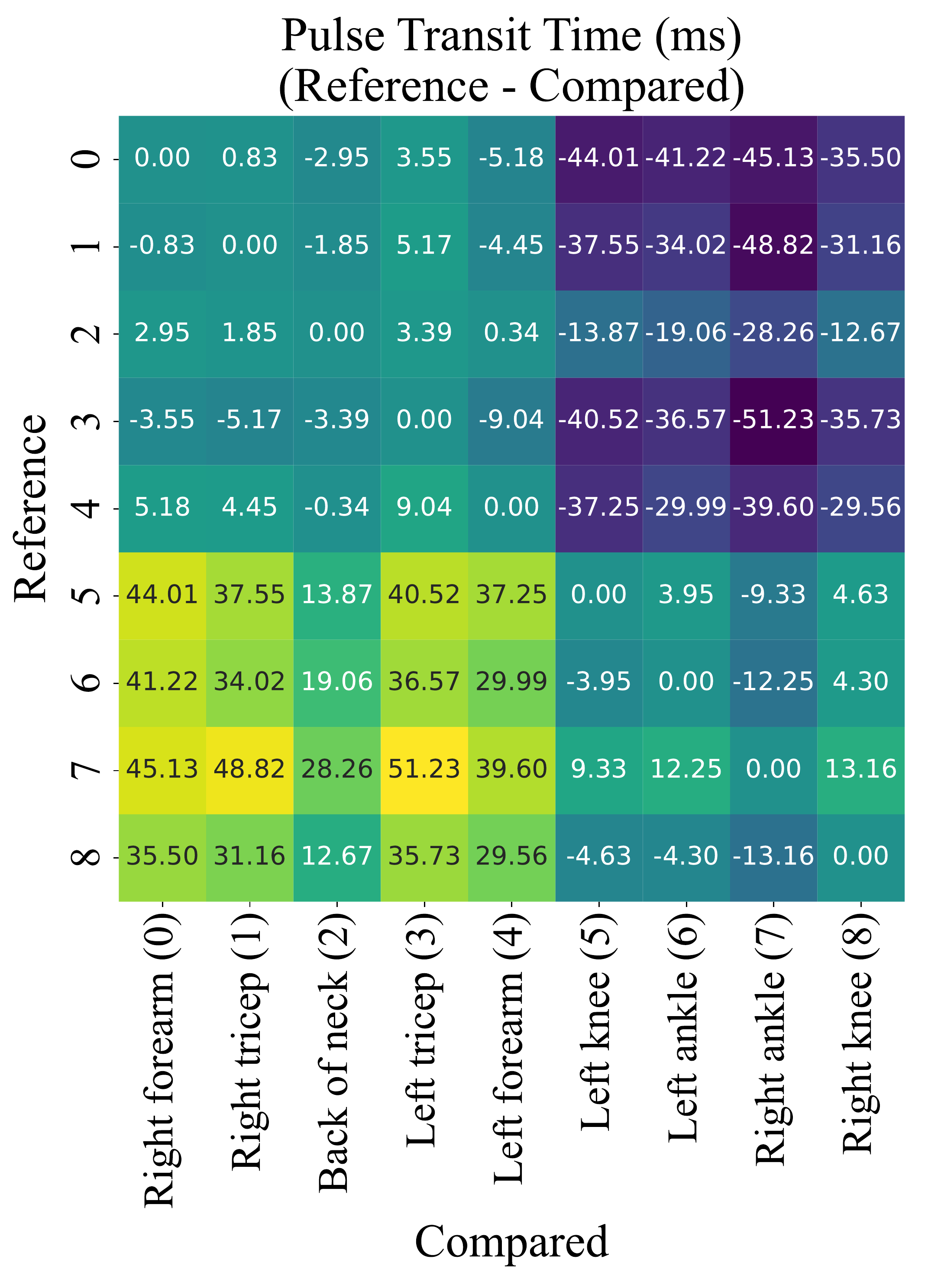}
        \caption{}
        \label{subfig:ptts}
        \end{subfigure}
        \end{minipage}
        \hfil
        \begin{minipage}[t]{0.25\linewidth}
        \begin{subfigure}{\linewidth}
        \includegraphics[width=\linewidth]{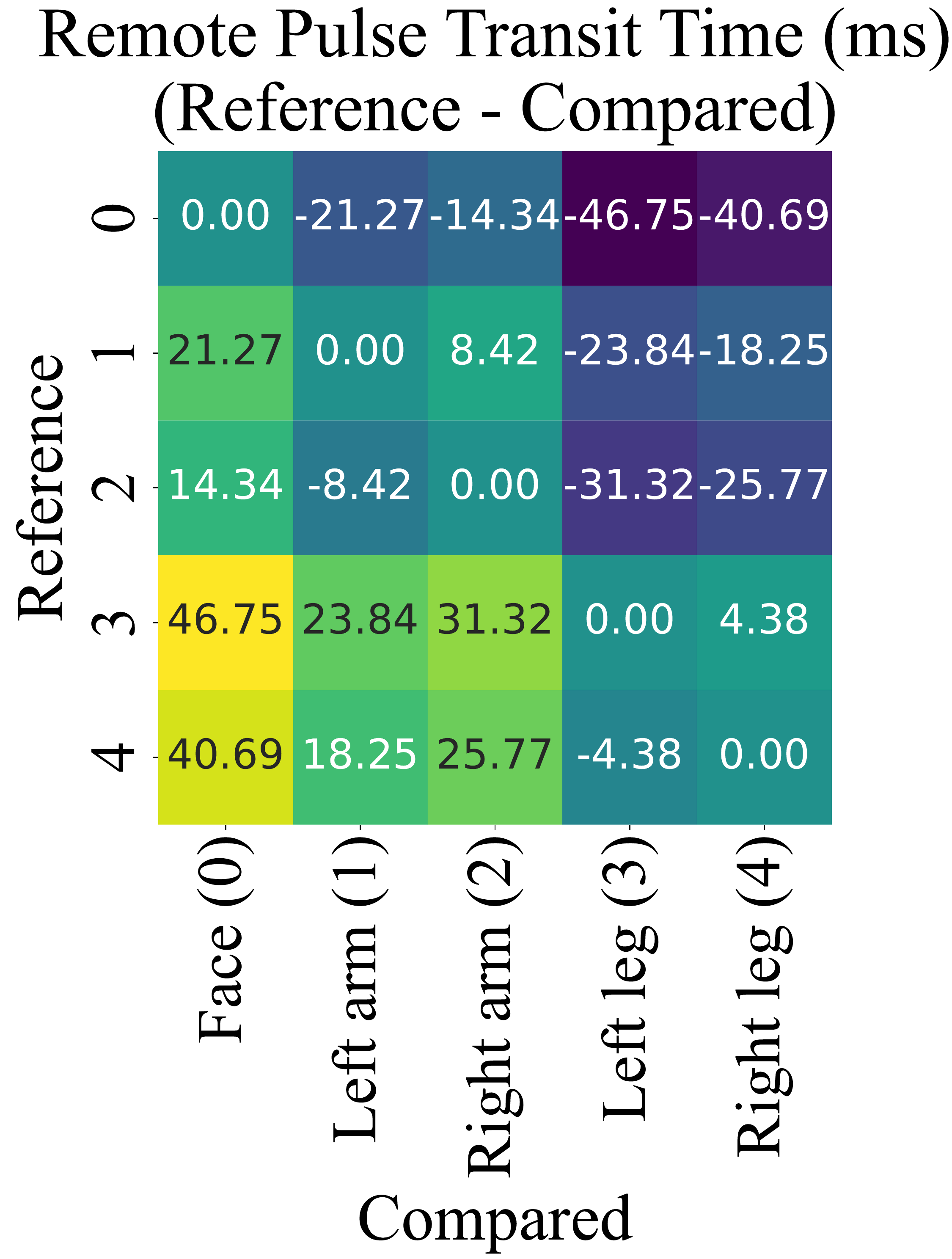}
        \vspace{7.27em}
        \caption{}
        \label{subfig:rptts}
        \end{subfigure}
        \end{minipage}
        \hfil
        \begin{minipage}[t]{0.37\linewidth}
        \begin{subfigure}{\linewidth}
        \includegraphics[width=\linewidth]{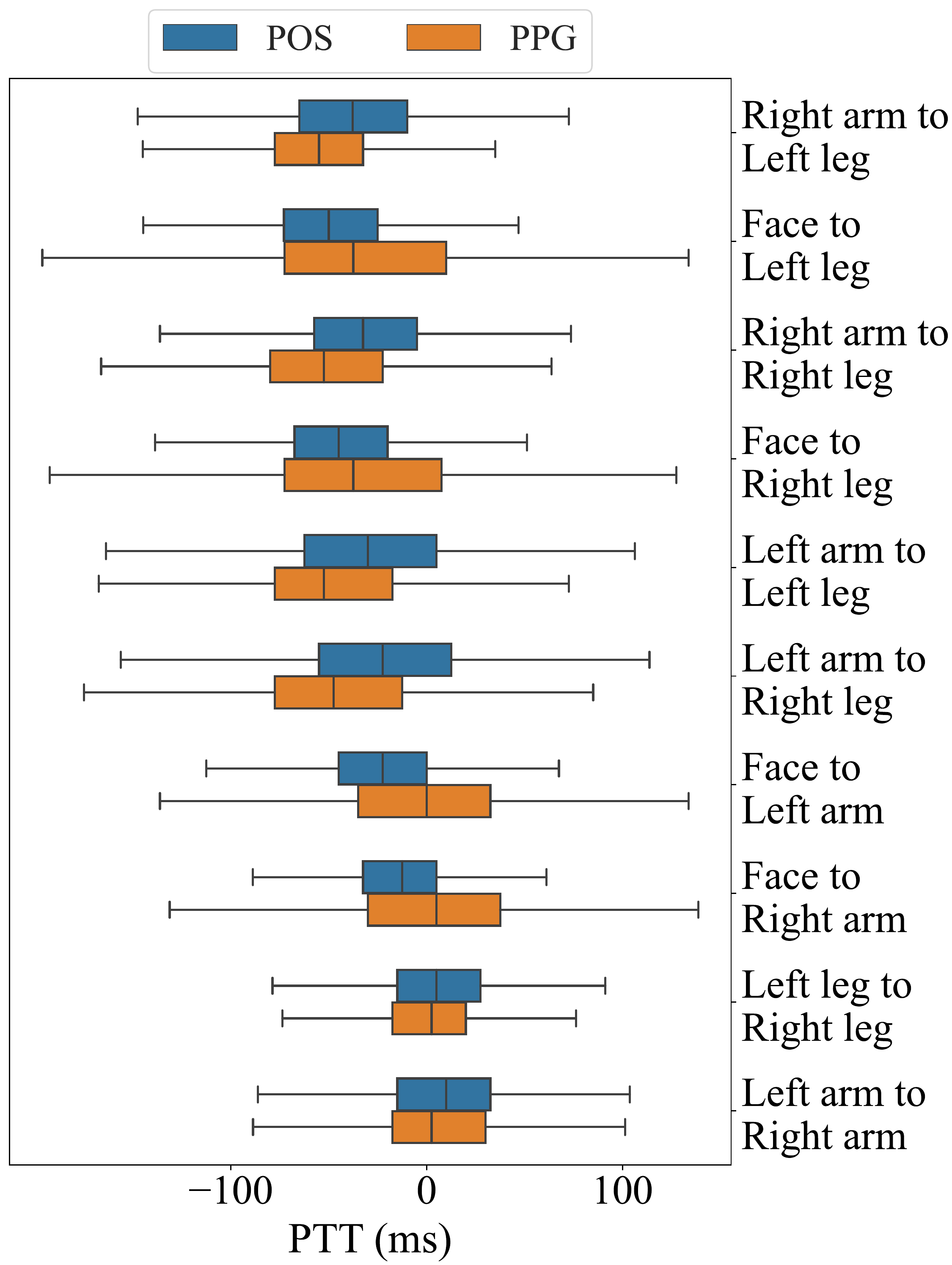}
        \caption{}
        \label{subfig:ptt_boxplots}
        \end{subfigure}
        \end{minipage}
    \caption{Pulse transit time from both contact PPG and rPPG calculated by a sliding cross-correlation between pairs of waveforms. (a) Time lags between the MAX30101 contact PPG sensors. (b) Time lags between POS waveform predictions on the five different body regions. (c) Grouped boxplots comparing the pulse transit time from PPG and rPPG. For the PPG measurements on the arms and legs we only used the forearm and knee sensors, respectively. The PTT is calculated as $X-Y$ given a label of measurement sites ``$X$ to $Y$".}
    \label{fig:ptts}
\end{figure*}

\subsection{Pulse Transit Time for Contact PPG Sensors}
Contact PPG measurements at different sites on the body contain small phase offsets due to different proximities to the heart. These offsets are sometimes referred to as differential pulse transit times (dPTT)~\cite{Block2020} and are negatively correlated with blood pressure. Throughout the paper we treat PTT and dPTT interchangeably. When fusing the PPG measurements at multiple sites into a single ground truth, we summed the waveforms without applying a phase shift. To justify this simplification, we calculated the differential pulse transit time between all pairs of sensors. Fig. \ref{subfig:ptts} shows a heatmap of the phase offsets between all pairs of PPG signals in milliseconds.

The phase differences were calculated with a sliding cross-correlation. We used a window size of 5 seconds (2,000 points) with a stride of 10 milliseconds (4 points), and a maximum lag 300 milliseconds, which is much higher than a typical transit time~\cite{Block2020}.
The sliding cross-correlation approach for PTT analysis is simple, but could be improved in future work by using foot-finding methods that measure time differences at the diastolic foot~\cite{Mukkamala2015}. 
The index with the maximum correlation was selected as the phase shift for each window. All pairs of transit times formed a skew-symmetric matrix, where $A^T = -A$.

In general, the phase offset between different sites is quite small relative to the pulse rate frequency. For example, the largest average phase offset of 51 milliseconds occurs between the left tricep and the right ankle. Given even a high pulse rate of 180 bpm, the phase angle for such a shift is only 27.66 degrees. Since most phase offsets are lower, there is very little risk of interference when summing the waveforms. The pulse transit times provide interesting physiological measurements of blood flow throughout the body. As mentioned previously, the largest difference occurs between the triceps and ankle. The triceps receive blood faster via the brachial artery and a closer proximity to the aorta than the lower legs. We had originally theorized that the neck sensor would sense the pulse wave before the other sensors, but our analysis refutes this.

\subsection{Pulse Transit Time from rPPG}
Beyond the benefit of contactless monitoring, one of the most powerful properties of a camera is the ability to sense at multiple spatial locations. To leverage this, we performed remote pulse transit time (rPTT) with a sliding cross-correlation from the POS rPPG waveforms at multiple sites on the body. The POS waveforms are noisier than the PPG waveforms in the legs and arms, but the infrequent pulse rate errors (see Fig. \ref{fig:result_HRs}) indicate that the signal quality is high enough for rPTT.

Figure \ref{subfig:rptts} shows a heatmap of the rPTT between all pairs of the face, legs, and arms. The maximum observed rPTT is 46.75 ms between the face and left leg, which is within a reasonable range compared to the largest contact-PTT measurements between triceps and legs (51.23 ms and 48.82 ms). An interesting observation is the bilateral asymmetry visible in the rPTT left and right sides of the body. The pulse waves from the right side of the body were consistently observed sooner than the left side of the body.

To compare the rPTT measurements with the PPG-derived PTT values we selected the nearest contact sensors to the rPTT measurement sites, and discarded the rest (\eg ankles and triceps). Figure \ref{subfig:ptt_boxplots} shows boxplots of the PTT values for all 5-second time windows in the relaxation portion of the dataset. The groups were sorted by the median PTT at the pair of measurement sites for easier viewing. The means of the rPTT measurements via POS are not perfectly aligned with the PTTs --- but upon further inspection, we can see that rPTT and PTT correlate well across the pairs of body sites. A potential reason for the bias in the rPTT values is that the signal quality is not uniform over the measured body sites. For example, the POS signal on the arms is likely shifted in phase towards the pulse wave at the hands. Similarly, the POS wave is likely shifted more towards the phase of the thighs than the lower legs. Furthermore, the sampling rate of the video is 90 fps, so large time lags such as 55 ms will only occur as 5 frames. Future work will perform more robust transit time calculations via the systolic foot and finer-grained spatial measurements of rPTT to better align with the contact-PPG sensors.

\section{Conclusion}
This paper offers a new Multi-Site Physiological Monitoring
(MSPM) dataset with multiple contact-based PPG measurements and simultaneous video to enable rPPG from multiple parts of the human body including the face, arms, legs, and neck. We applied various rPPG methods, including CHROM\cite{DeHaan2013}, POS\cite{Wang2017} and RPNet\cite{Speth_IJCB_2021}, to MSPM samples and performed pulse rate estimation on all face and non-face body regions. We observed that the heart rate estimation from non-face body parts, especially palm and arms, showed a promising accuracy compared to face-based rPPG. However, the deep learning-based (3D-CNN) approach evaluated in this paper, and trained entirely on facial data, learned spatial pulse estimation priors which disrupted its ability to generalize to non-facial regions. This suggests that training deep learning rPPG systems using non-facial regions may be needed to offer good fidelity for rPPG outside the face region. We also positively verified a possibility to apply rPPG to estimate pulse transit times (PTT) between various body parts by comparing rPPG-based PTT with PPG-based PTT. We believe that these baseline experiments along with the new MSPM dataset opens an interesting research area focused on non-face multi-site rPPG.

\section*{Acknowledgements}
This research was sponsored by the Securiport Global Innovation Cell, a division of Securiport LLC. Commercial equipment is identified in this work in order to adequately specify or describe the subject matter. In no case does such identification imply recommendation or endorsement by Securiport LLC, nor does it imply that the equipment identified is necessarily the best available for this purpose. The opinions, findings, and conclusions or recommendations expressed in this publication are those of the authors and do not necessarily reflect the views of our sponsors.

{\small
\bibliographystyle{ieee_fullname}
\bibliography{egbib}
}

\end{document}